# Exploring Semantic Segmentation on the DCT Representation

Shao-Yuan Lo    Hsueh-Ming Hang
National Chiao Tung University
sylo95.eecs02@g2.nctu.edu.tw, hmhang@nctu.edu.tw

## ABSTRACT

Typical convolutional networks are trained and conducted on RGB images. However, images are often compressed for memory savings and efficient transmission in real-world applications. In this paper, we explore methods for performing semantic segmentation on the discrete cosine transform (DCT) representation defined by the JPEG standard. We first rearrange the DCT coefficients to form a preferred input type, then we tailor an existing network to the DCT inputs. The proposed method has an accuracy close to the RGB model at about the same network complexity. Moreover, we investigate the impact of selecting different DCT components on segmentation performance. With a proper selection, one can achieve the same level accuracy using only 36% of the DCT coefficients. We further show the robustness of our method under the quantization errors. To our knowledge, this paper is the first to explore semantic segmentation on the DCT representation.

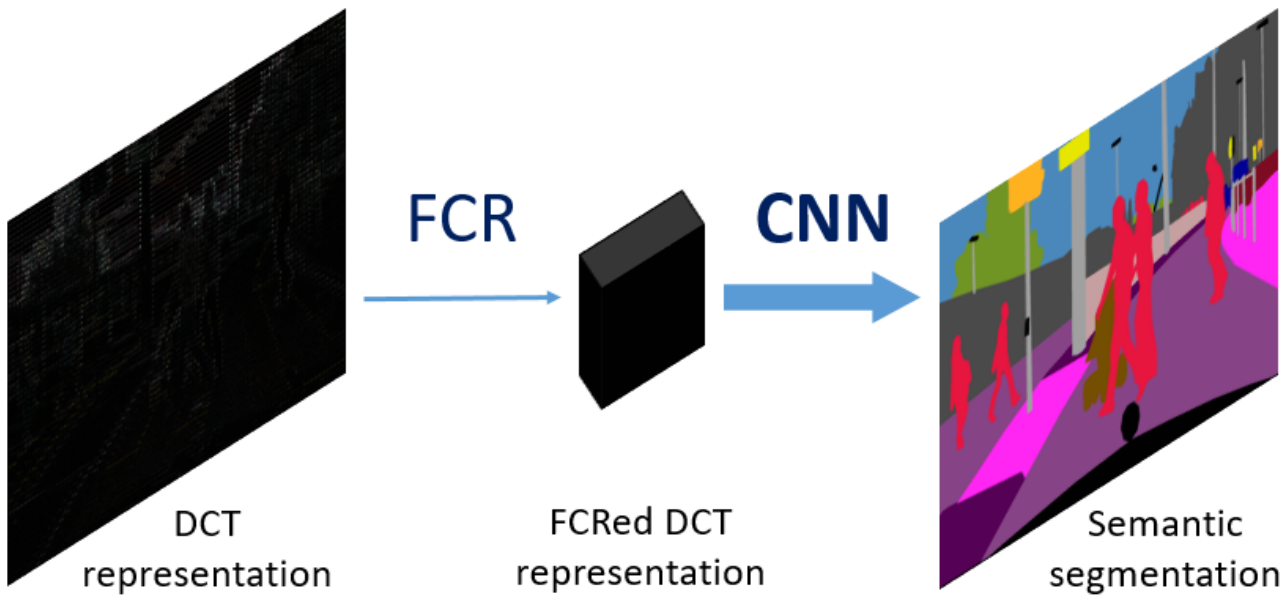

**Figure 1: Flowchart of the proposed method of semantic segmentation on the DCT representation.**

## 1 INTRODUCTION

Deep convolutional neural networks (CNNs) have shown tremendous success in a variety of computer vision tasks, such as image classification [10,13], object detection [6,15], and semantic segmentation [1,16,17]. These CNN models have been broadly adopted not only in academia but also in industry. In many real-world applications, images/videos are compressed into specific coding formats for storage savings and high-speed transmission. For example, JPEG [24], GIF, PNG are the widespread image compression standards, and H.264, H.265 (HEVC) are the recent video compression standards. However, most of the existing CNNs are trained to perform on the RGB format images, i.e., the input data are expressed as an array of RGB pixels. Hence, the compressed data have to undergo a decompression step before being fed to a CNN. However, this step is time-consuming and requires high computation and memory demands, so it is favored to skip it. Therefore, performing computer vision tasks in the compressed domain has become an emerging research topic.

The discrete cosine transform (DCT) representation encodes the original spatial-domain RGB images into components in the frequency domain. The DCT representation has been widely used in image compression, and it is the core processing of the JPEG compression standard (with a block size of 8×8). The JPEG standard was established in 1992, but it is still one of the most popular image file formats today. Except for JPEG 2000, the recent image/video coding standards (H.264 and H.265) also adopt DCT (with possibly different sizes). Several researchers started looking into the possibilities of using DCT coefficients to do the classification or detection tasks. In this study, we extend this direction to semantic segmentation, exploring the semantic segmentation task in the DCT domain.

In this paper, we propose a method for conducting semantic segmentation on the compressed DCT representation. Figure 1 shows the pipeline. We first rearrange the order and format of the DCT coefficients before feeding them to CNNs. After the rearrangement, the spatial relationship of these coefficients are represented on the first and the second dimensions (spatial dimension), and their frequency representation is coded on the third dimension. We call this technique as Frequency Component Rearrangement (FCR). Next, we choose EDANet [16] as our baseline network owing to its good balance between performance and complexity. Because the spatial resolution of the FCRed DCT representation is 1/8 to its RGB counterpart, the three downsampling operations of EDANet would make the feature size too small. Therefore, we modify EDANet by removing all the downsampling operations to facilitate the DCT input data size. We call the modified network as DCT-EDANet. The complexity reduction due to the smaller input spatial size enables DCT-EDANet to be deeper, increasing its capacity. Our method achieves an accuracy close to that of the RGB baseline at about the same network complexity. To the best of our knowledge, we are the first to explore this research topic, DCT-domain semantic segmentation.

One step further, we examine how different DCT components would affect performance. We select different numbers of low-frequency sub-bands from each of the three Y, Cb, and Cr channels (YCbCr color space), and form various combinations of the coefficients as the system inputs for both training and testing. We demonstrate the impact of these components on performance, and identify the best input coefficient proportion of the Y, Cb, and Cr components, which can achieve a similar accuracy using only 36% of input coefficients compared to the model with 100% input coefficients. The effect of quantization using different quality factors is also investigated. The proposed system is robust against serious quantization errors.

The rest of this article is organized as follows. Section 2 covers several related work on the compressed-domain computer vision. Section 3 reviews the JPEG compression, DCT operation, and DCT representation. In Section 4, we describe the proposed method and its implementation results. Then, Section 5 and Section 6 report our experiments on the coefficient selection and quantization effect, respectively. Finally, Section 7 concludes this study.

## 2 RELATED WORK

Several studies on compressed-domain analytics have been reported using the conventional image processing techniques. These studies aim at, image indexing [18], face recognition [8], image retrieval [9], etc. In the deep learning age, a few researchers started employing CNNs to do analytics in the compressed-domain. One example is multiple JPEG compression detection. Verma *et al.* [23] computed histograms of different DCT sub-bands, then concatenated these histograms to form a 1-D vector. This vector would be sent to a 1-D CNN to do detection. The multi-branch CNN architecture [14] processes different DCT sub-bands separately, then these features are concatenated for the final inference.

Another research trend is to develop tailored CNNs to DCT input for image classification or object detection. Ghosh and Chellappa [5] applied the DCT operation to the feature maps generated by the first convolutional layer of their CNN, accelerating the convergence in training. Ulicny and Dahyot [20] used a CNN to classify images in the DCT domain. Ulicny *et al*. [21,22] designed the so-called harmonic blocks to replace the traditional convolutions. These blocks generate features by learning combinations of spectral filters defined by DCT. Gueguen *et al*. [7] developed ResNet-50 [10] variants, which can operate on JPEG data. It is shown to be faster and more accurate. Ehrlich and Davis [4] created a DCT-domain ResNet, which is mathematically equivalent to the spatial-domain network, by including the transform into network weights. SSD_freq [3] is a modified version of SSD [15] so that it is able to process the DCT inputs. It is a pioneer in the JPEG-domain object detection.

Moving towards a different direction, Torfason *et al*. [19] focused on the learned compressed representation other than JPEG. They jointly trained a compression network with an inference network and bring performance gain. On the video side, Wu *et al*. [25] designed a compressed video action recognition system by using separate networks for I-frames and P-frames. Their approach is more efficient than the conventional 3D convolution structures.

## 3 JPEG COMPRESSION

This work is motivated by the JPEG standard [24], a dominant image file format. The JPEG compression algorithm is summarized in the following steps.

1. Convert the color space from RGB to YCbCr.
2. Perform block-wise (8×8 pixels) DCT.
3. Quantize the DCT coefficients by a quantization matrix.
4. Encode the coefficients by entropy encoding.

The YCbCr color space consists of a *luma* (luminance) component (Y), and two *chroma* (chrominance) components (Cb and Cr). The *chroma* channels are usually subsampled by a factor of 2, for human vision is less sensitive to subtle color changes than to subtle illuminance changes. In order to observe the influence of different DCT coefficients on performance, we bypass the *chroma* subsampling step for the DCT representation in our implementation. Each of the three channels is partitioned into blocks of 8×8 pixels, and 128 is subtracted from all the pixel values. Then, each block is transformed by 2-D DCT (type-II), which is defined as:

$$G_{u,v} = \frac{1}{4}\alpha_u \alpha_v \sum_{x=0}^{7}\sum_{y=0}^{7} g_{x,y} \cos[\frac{(2x+1)u\pi}{16}] \cos[\frac{(2y+1)v\pi}{16}] \qquad (1)$$

where $\alpha_u$ and $\alpha_v$ are the normalizing factors, $g_{x,y}$ is the pixel value at $(x,y)$, $G_{u,v}$ is the DCT coefficient at $(u,v)$, and $0 \leq u, v < 8$.

In the DCT domain, the pixel information is represented by spatial frequency spectrums. The upper-left of each 8×8 block comprises low-frequency sub-bands, while the high-frequency sub-bands are located on the bottom-right. If the compression is lossy, the frequency coefficients are quantized by a quantization matrix and rounded to integers. Because human eyes are less sensitive to high-frequency variations, the high-frequency sub-hands are recorded with a lower accuracy or even discarded. Finally, the quantized coefficients are coded by the run-length encoding (RLE) and Huffman coding. Since the structure of Huffman codes is incompatible with the input of CNNs, we follow the typical compressed-domain analytic setup; that is, use the outputs of Step 2 or Step 3 as our DCT representation.

## 4 METHOD

We develop a method for doing semantic segmentation on the DCT representation. In this section, we first describe the Frequency Component Rearrangement (FCR) technique, which is used to rearrange the DCT coefficients to make them be easily exploited by CNNs. Next, we introduce the proposed DCT-EDANet, a modification of EDANet [16] to operate in the DCT domain. Finally, the benchmark, training procedure, and experimental results are reported.

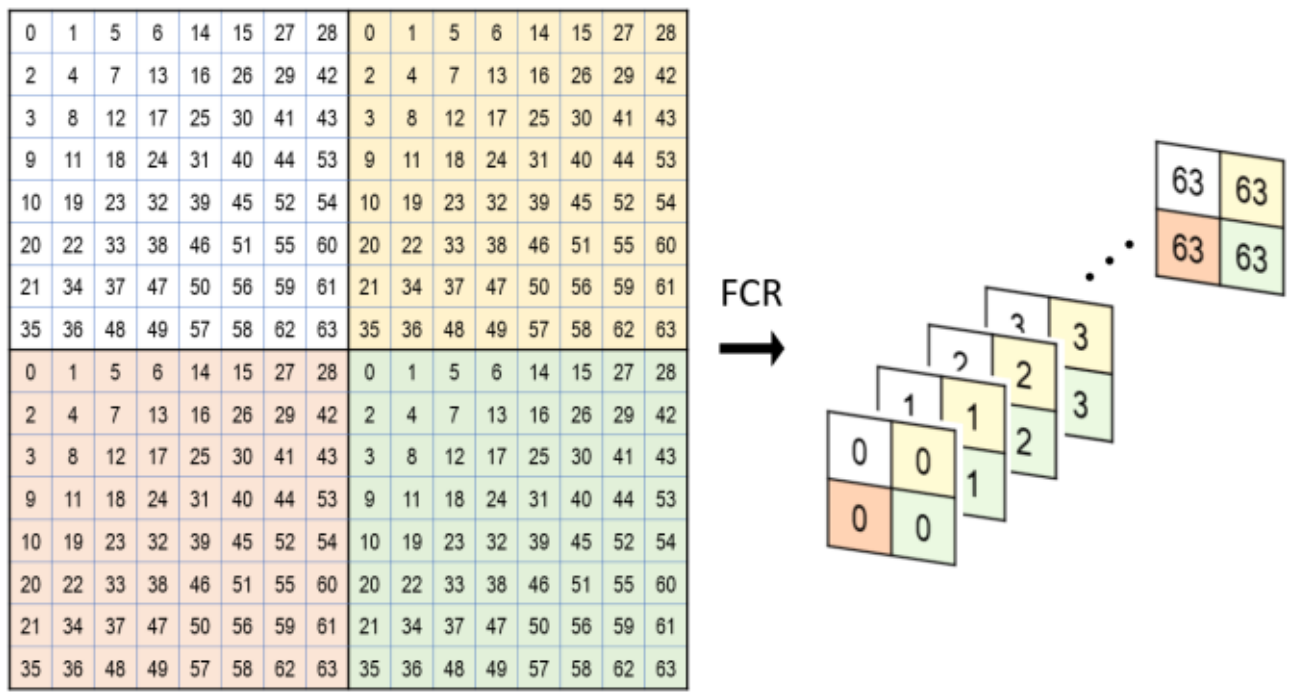

**Figure 2: Illustration of FCR. It takes an image with size (16, 16, 1). The left side is the outputs of DCT. The right side is the FCRed DCT representation (64 channels).**

## 4.1 Frequency Component Rearrangement

After performing block-wise DCT on each of the Y, Cb, and Cr channels, the spatial dimension (each channel) contains not only the spatial relationship but also the frequency relationship of the DCT coefficients. To be more specific, in an 8×8 block, the location of each coefficient in this block corresponds its frequency index, i.e., each coefficient represents a specific frequency component. However, when a CNN performs the convolution operation, such frequency relationship is treated as their spatial relationship. This makes CNNs misinterpret the input information and fail to extract essential features in the DCT domain.

To solve this problem, we use FCR to rearrange the DCT coefficients before feeding them into CNNs. FCR reshapes each block of dimension (8, 8, 1) into (1, 1, 64). That is, each frequency component is placed at its corresponding channels on the third dimension. For an RGB image with a size of ($h$, $w$, 3), the size of its FCRed DCT representation is ($h/8$, $w/8$, 192), where 192 comes from 3 × (8 × 8). Figure 2 gives an example of $h$ = $w$ = 16. With FCR, the spatial dimension contains simply the spatial relationship of these coefficients, and their frequency relationship is purely represented on the third dimension. In this manner, the 2-D convolution is applied to the spatial neighbors of the same frequency component, so that CNNs are able to exploit the DCT representation properly.

## 4.2 Network Design

In general, a CNN includes several downsampling operations to integrate the input information and enlarge the receptive field. The selected baseline network, EDANet, contains three downsampling operations (see Figure 3a). In other words, the size of feature maps at the final stage (EDA Block 2) is 1/8 to the input images. As discussed in Section 4.1, the spatial resolution of the FCRed DCT representation is 1/8 of its RGB counterpart. Thus, the downsampling operations would results in a feature map size of 1/64 of the RGB image, which is too small. Particularly, the spatial information and boundary details are important to localize objects in semantic segmentation.

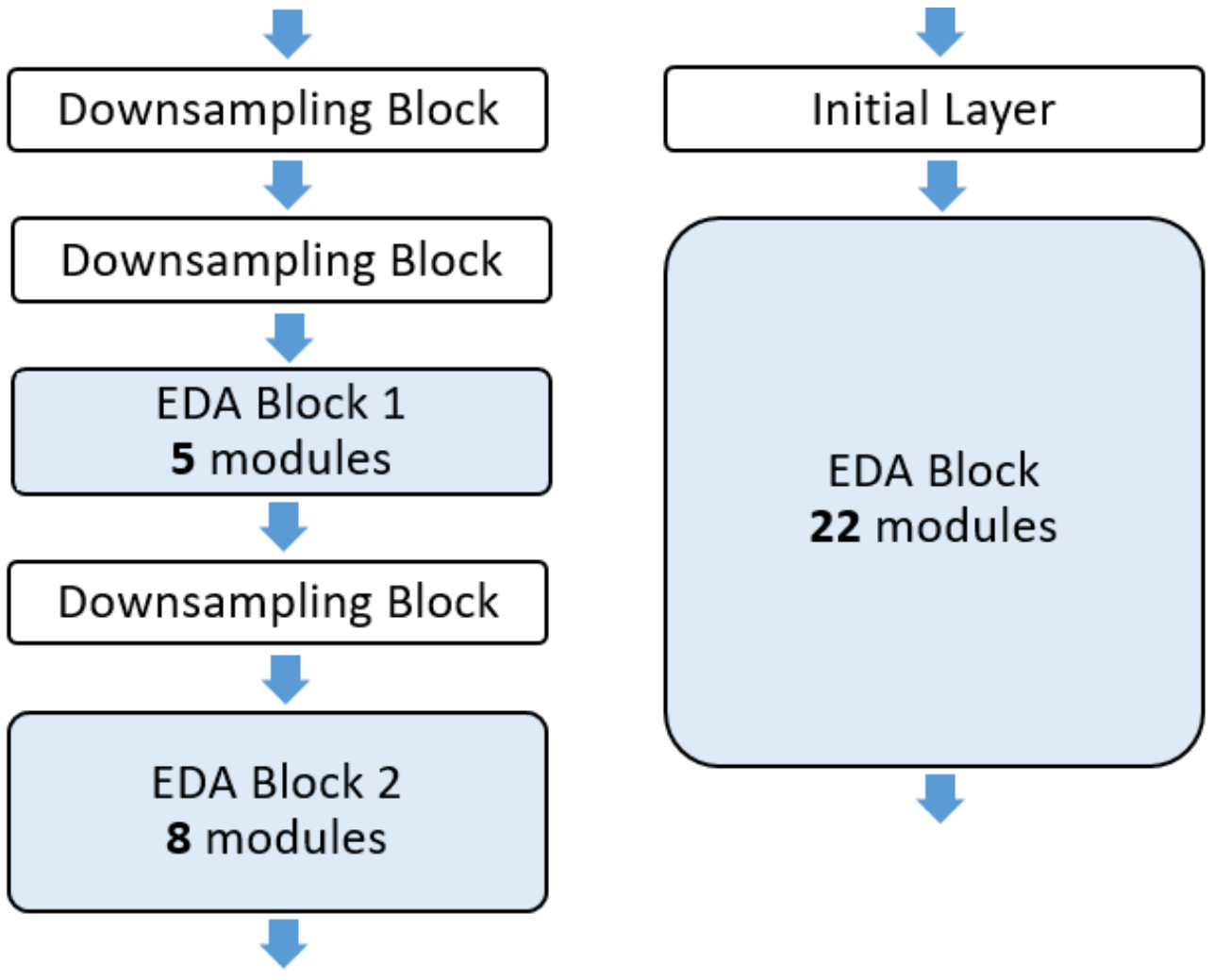

(a) (b)

**Figure 3: The structures of (a) the baseline network, EDANet [16] and (b) the proposed DCT-EDANet.**

The proposed DCT-EDANet is modified from EDANet mainly on two aspects, removing downsampling and increasing depth. Figure 3b shows its structure. We remove all the downsampling operations to accommodate our target image representation. In this way, the output feature maps maintain the size of 1/8 to the original RGB image, and thus the necessary spatial information remains. The Initial Layer is a vanilla 3×3 convolutional layer with batch normalization [11] and ReLU. It is similar to the Dowsampling Block of EDANet but has no downsampling operation. In addition, because the entire DCT-EDANet operates on a spatial resolution of 1/8, the computational complexity is significantly reduced. As a result, we can substantially deepen DCT-EDANet. At the same level of computational complexity of EDANet, it can include 22 EDA modules in total.

## 4.3 Benchmark and Training Procedure

We use the Cityscapes dataset [2] as the benchmark. It is a road scene dataset for the semantic segmentation task. The dataset has 19 object classes, and it consists of 2975, 500, and 1525 images for training, validation, and testing, respectively. Its original image resolution is 1024×2048. Since EDANet is our baseline, we conform to the setup of EDANet, which trains and tests networks on the downsampled 512×1024 images.

We follow the training setup similar to that in EDANet. Our networks are trained by using the Adam optimizer [12] with batch size 10 and weight decay $1e^{-4}$. The poly learning rate policy is used, which multiplies the learning rate by $(1 - iter/\max_iter)^{power}$ with power 0.9. We set the initial learning rate to $5e^{-4}$. A class weighting scheme, $w_{class} = 1/log(p_{class} + k)$, is included, where $k$ is set to 1.12. We adopt random horizontal flip and the translation of 0~2 pixels on both axes for data augmentation. All models are evaluated by the mean of intersection-over-union (mIoU) metric. We do not include any testing tricks, such as multi-crop and multi-scale testing. In our experiments, we train models with one-stage and through just 2/3 number of iterations compared to that in EDANet [16] since we compare the relative accuracy in our analysis. Our experiments are conducted on a single GTX 1080Ti.

### 4.4 Results

We evaluate the performance of the proposed DCT-EDANet on the Cityscapes dataset. We also compare it to the models of the original EDANet architecture with different types of input representation. These models are as follows. EDANet-RGB uses the RGB input, EDANet-DCT uses the original DCT input, and EDANet-FCRedDCT uses the FCRed DCT input. Table 1 reports the experimental results. EDANet-RGB serves as our baseline. Since the purpose of ablation experiments here is to verify our approach, training the models through 2/3 number of iterations compared to the implementation of the authors of EDANet is enough. Hence, the mIoU accuracy of EDANet-RGB is slightly lower than that reported in the EDANet paper [16]. Next, EDANet-DCT is obviously less accurate than EDANet-RGB though they have identical computational cost and an identical amount of input information. This reflects the issue that EDANet is not tailored to the DCT input. When the FCR technique is adopted, the performance becomes even worse owing to the extremely small feature size. Compared to EDANet-FCRedDCT, the proposed DCT-EDANet obtains a dramatic improvement, showing the importance of larger feature size.

Moreover, DCT-EDANet surpasses EDANet-DCT at about the same network complexity. A similar result is observed in SSD_freq [3]. SSD_freq does not employ FCR, which leads to lower accuracy than its RGB counterpart. By contrast, other studies that adopt ideas similar to FCR perform better [7,14]. These results demonstrate the effectiveness of FCR. On the other hand, Table 2 views this issue from another perspective. EDANet-DCT-1/4coef and DCT-EDANet-1/4coef are EDANet-DCT and DCT-EDANet that use 1/4 number of total DCT coefficients as inputs, respectively. More precisely, they take the first 16 low-frequency coefficients from each 8×8 block. In this manner, EDANet-DCT-1/4coef has an input size of (*h/2*, *w/2*, 3), i.e., each 8×8 block is condensed to 4×4. DCT-EDANet-1/4coef, who adopts the FCR, has an input size of (*h/8*, *w/8*, 48). Under this situation, the accuracy gap between these two models is widened from 2.3% to 4.0%, which indicates DCT-EDANet are more favorable when the inputs are condensed. A detailed investigation in this issue is discussed in Section 5. In summary, when the spatial relationship and the frequency relationship are encoded on separate dimensions, CNNs achieve better performance. Figure 4 shows some visual results.

Still, DCT-EDANet is narrowly defeated by EDANet-RGB. The reason is that spatial information is particularly paramount to semantic segmentation, but the DCT format reduces the spatial resolution in trading of the frequency decomposition. It is thus more challenging to do semantic segmentation on the DCT representation.

**Table 1: Evaluation results of the proposed method and other models. Multi-Adds: the number of multiply-add operations.**

| Architecture | Input | mIoU (%) | Multi-Adds |
|---|---|---|---|
| EDANet | RGB | 63.7 | 8.97B |
| EDANet | DCT | 59.3 | 8.97B |
| EDANet | FCRed DCT | 37.8 | 0.20B |
| DCT-EDANet | FCRed DCT | 61.6 | 8.52B |

**Table 2: Comparison between EDANet-DCT and DCT-EDANet using 1/4 DCT coefficients as input.**

| Model | mIoU (%) |
|---|---|
| EDANet-DCT-1/4coef | 55.0 |
| DCT-EDANet-1/4coef | 59.0 |

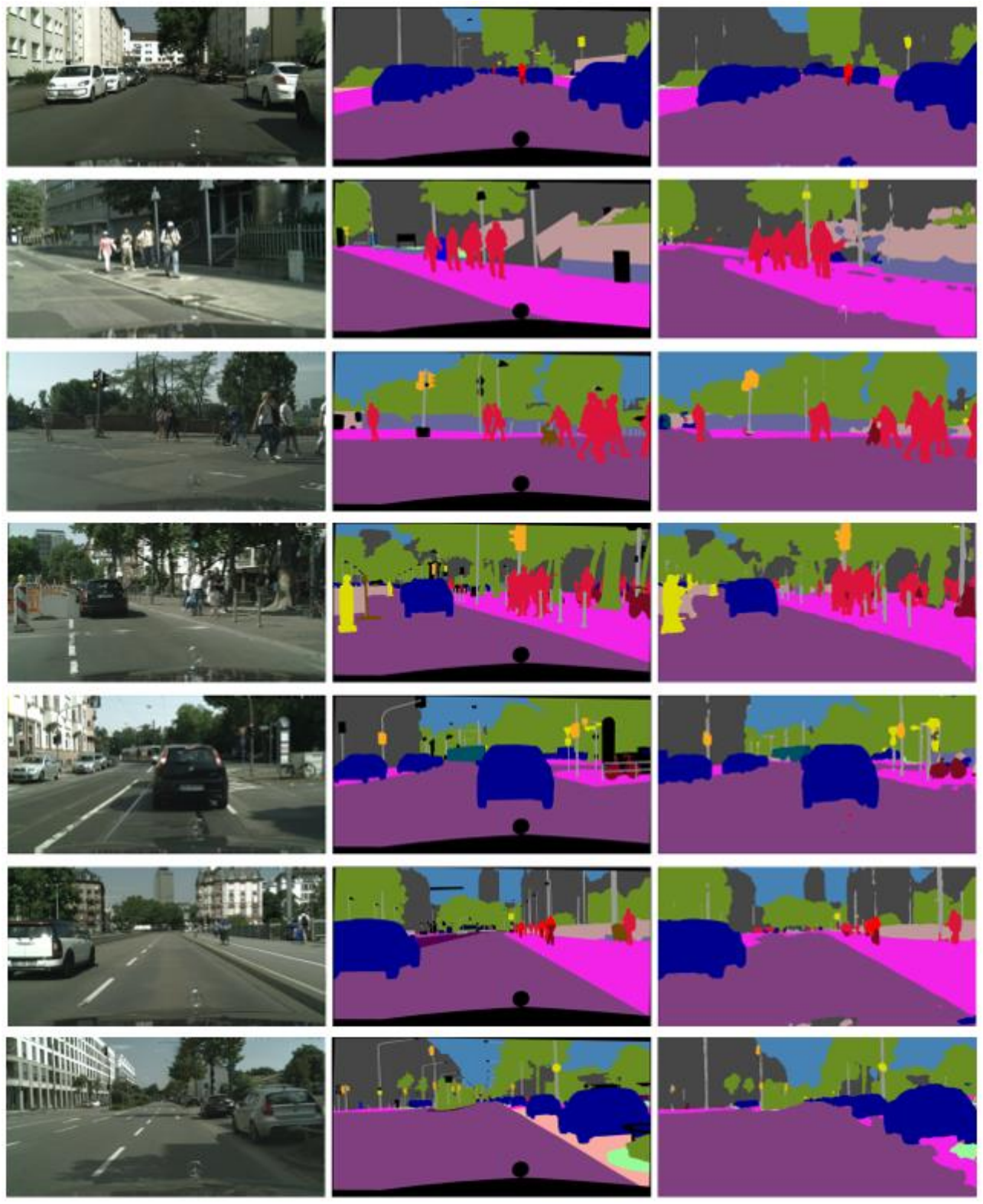

**Figure 4: Sample results of the proposed DCT-EDANet on the Cityscapes validation set. From left to right: (a) RGB images, (b) Ground truths, (c) DCT-EDANet.**

Object detection is another task focusing on localization. Therefore, SSD_freq, the first method for detection in the DCT domain, still has a lower performance than its RGB counterpart by a huge margin. Compared to SSD_freq, our method is very close to its baseline.

**Table 3: Experimental results on frequency component selection. Each model is both trained and tested by its listed numbers of low-frequency components.**

| Model | # input components | # Y components | # Cb components | # Cr components | mIoU (%) |
|---|---|---|---|---|---|
| DCT-EDANet | 192 | 64 | 64 | 64 | 61.6 |
| M-64-0-0 | 64 | 64 | 0 | 0 | 59.8 |
| M-49-9-9 | 67 | 49 | 9 | 9 | 60.6 |
| M-36-16-16 | 68 | 36 | 16 | 16 | 61.2 |
| M-25-25-25 | 75 | 25 | 25 | 25 | 59.7 |
| DCT-EDANet-1/4coef | 48 | 16 | 16 | 16 | 59.0 |
| M-16-4-4 | 24 | 16 | 4 | 4 | 59.9 |
| M-16-1-1 | 18 | 16 | 1 | 1 | 57.4 |
| M-9-4-4 | 17 | 9 | 4 | 4 | 58.7 |
| M-0-0-16 | 16 | 0 | 0 | 16 | 46.4 |

**Algorithm 1** JPEG Quantization Algorithm

**Require:** Base quantization matrix $Qb$

$$Qb = \begin{bmatrix} 16 & 11 & 10 & 16 & 24 & 40 & 51 & 61 \\ 12 & 12 & 14 & 19 & 26 & 58 & 60 & 55 \\ 14 & 13 & 16 & 24 & 40 & 57 & 69 & 56 \\ 14 & 17 & 22 & 29 & 51 & 87 & 80 & 62 \\ 18 & 22 & 37 & 56 & 68 & 109 & 103 & 77 \\ 24 & 35 & 55 & 64 & 81 & 104 & 113 & 92 \\ 49 & 64 & 78 & 87 & 103 & 121 & 120 & 101 \\ 72 & 92 & 95 & 98 & 112 & 100 & 103 & 99 \end{bmatrix}$$

Quality factor, $qf$
Multiplier, $m$
Objective quantization matrix, $Q$
Unquantized DCT coefficient block, $U$
Quantized DCT coefficient block, $V$
**if** $qf < 50$
  $m = 5000 / qf$
**else**
  $m = 200 - 2qf$
**end**
$Q$ = **floor** [ $(Qb \times m + 50) / 100$ ]
**if** $Q_{i,j} = 0$ **for** $i = 0, 1, \ldots, 7$; $j = 0, 1, \ldots, 7$
  $Q_{i,j} = 1$
**end**
$V_{i,j}$ = **round** ( $U_{i,j} / Q_{i,j}$ ) **for** $i = 0, 1, \ldots, 7$; $j = 0, 1, \ldots, 7$

**Table 4: Experimental results on quantization.**

| Model | Quality factor | mIoU (%) |
|---|---|---|
| DCT-EDANet | No | 61.6 |
| M-QF70 | 70 | 60.5 |
| M-QF50 | 50 | 60.6 |
| M-QF30 | 30 | 60.0 |

## 5 FREQUENCY COMPONENT SELECTION

In this section, we investigate the impact of different combinations of DCT components on segmentation performance. We select different numbers of low-frequency coefficients from each of the Y, Cb, and Cr channels to form various combinations of these components. Then, each combination is used as the inputs to the networks for both training and testing stages. Table 3 reports our experimental results.

From M-64-0-0 to M-25-25-25, each model has a near number of input components that ranges between 64 and 75. We can see that M-36-16-16 performs the best, M-49-9-9 is the runner-up, and M-25-25-25 is the worst. M-36-16-16 attains an accuracy close to that of DCT-EDANet (with full input components) by using barely 36% input components. This indicates that the best proportion of the Y, Cb and Cr coefficients is approximately 50:25:25. That is, the Y information plays the most important role, but a certain amount of the Cb and Cr is needed to support. This result is consistent with a principle of the JPEG compression algorithm, in which the *chroma* information is less critical and thus subsampled in the JPEG codec. M-49-9-9 has a similar proportion, so it gets a second. M-64-0-0 only has the Y components without the aid of the Cb and Cr, and M-25-25-25 uses an equal ratio without a sufficient amount of Y information. Therefore, they are the last.

From M-16-1-1 to M-0-0-16, they present a similar result. M-9-4-4 has the same ratio as M-36-16-16, and it is the winner within its group. In addition, we can also observe that M-16-4-4 outperforms DCT-EDANet-1/4coef, M-64-0-0, and M-25-25-25 by using significantly fewer input components, which also indicates that the proportion is critical. As a result, we find the best input component proportion is around 50:25:25, providing a guideline for future studies on DCT-domain analytics.

## 6 QUANTIZATION

In the JPEG codec, if the compression is lossy, the quantization step is included. Because human visual system is more sensitive to low-frequency variations, the resolution of high-frequency components can be reduced more significantly. The quality factor, ranging from 0 to 100, determines the degree of quantization. A small quality factor specifies a rough quantization, which means lower image quality but a smaller file size. The JPEG quantization procedure is described in Algorithm 1.

In this section, we investigate the influence of different quality factors on segmentation performance. Models M-QF70, M-QF50, and M-QF30 are trained and tested on the DCT coefficients quantized by their specified quality factor. The experimental results are shown in Table 4. We can see that even when the quality factor is as low as 30, the accuracy merely drops 1.6%. These results show that with heavy quantization, the compressed DCT components can still provide a comparable accuracy. It also demonstrates that the proposed method is able to tolerate serious quantization errors.

## 7 CONCLUSIONS

In this paper, we propose a solution for performing semantic segmentation on the DCT representation. We rearrange the DCT coefficients by using FCR. Then, we modify EDANet by discarding all the downsampling operations to maintain the DCT inputs at 1/8 spatial resolution, and by deepening the network to maintain the network capacity. We obtain near accuracy to the RGB baseline. Furthermore, we investigate the best proportion of YCbCr information selections, which can attain the same level of accuracy using significantly fewer coefficients. We also demonstrate our method is highly resistant to quantization errors, even when the quality factor is as low as 30. This work shows the feasibility of performing the challenging semantic segmentation task in the JPEG compressed domain. The elaborated analysis of DCT coefficient selections provides a guideline for future studies on compressed-domain analytics. Further improvements in performance and inference speed can be anticipated.

## ACKOWLEDGMENTS

We would like to thank Dr. Yen-Kuang Chen, Alibaba and Prof. Wen-Hsiao Peng, National Chiao Tung University, for their helpful comments and discussions.

This work is partially supported by the Ministry of Science and Technology, Taiwan under Grant MOST 108-2634-F-009-007 through Pervasive AI Research (PAIR) Labs, National Chiao Tung University.